\documentclass[letterpaper, 10 pt, conference]{ieeeconf}  

\IEEEoverridecommandlockouts                              

\usepackage{amsmath} 
\usepackage{amssymb}  
\usepackage{soul}
\usepackage{booktabs}
\usepackage{graphicx}
\usepackage{algorithm}
\usepackage{algorithmic}

\usepackage{amsthm}
\usepackage[dvipsnames]{xcolor}
\usepackage{subcaption}
\usepackage{multirow}
\usepackage{siunitx}
\usepackage{wasysym}  
\usepackage{colortbl}
\usepackage{pifont} 
\definecolor{goodbg}{RGB}{230,245,234} 

\newcommand{\req}{\textcolor{red!70!black}{\CIRCLE}}
\newcommand{\free}{\textcolor{green!40!black}{\Circle}}

\usepackage{balance}
\usepackage{cite} 
\usepackage{hyperref}
\usepackage{float}

\definecolor{best}{RGB}{200, 230, 201}
\definecolor{secondbest}{RGB}{225, 245, 214}

\title{\LARGE \bf
APEX: Action Priors Enable Efficient Exploration for Robust Motion Tracking on Legged Robots}

\author{
Shivam Sood$^{1}$, Laukik Nakhwa$^{1}$, Sun Ge$^{1}$, Yuhong Cao$^{1}$, Jin Cheng$^{2}$,
Fatemah Zargarbashi$^{2}$, Taerim Yoon$^{2,3}$, \\ Sungjoon Choi$^{3}$, Stelian Coros$^{2}$,
and Guillaume Sartoretti$^{1}$%
\thanks{$^{1}$Authors are with the Multi-Agent Robotic Motion Lab in the Department of Mechanical Engineering, College of Design and Engineering, National University of Singapore, Singapore.
        {\tt\small \{shivamsood, e1454338, sunge ,caoyuhong, guillaume.sartoretti\}@u.nus.edu}}%
\thanks{$^{2}$The authors are with the Computational Robotics Lab in the Department
of Computer Science, ETH Zurich, Switzerland
{\tt\small \{jicheng, fzargarbashi, scoros\}@ethz.ch}}%
\thanks{$^{3}$The authors are with the Robot Intelligence Lab in the Department
of Artificial Intelligence, Korea University, South Korea.
{\tt\small \{taerimyoon, sungjoon-choi\}@korea.ac.kr}}%
}

\begin{document}

\maketitle
\thispagestyle{empty}
\pagenumbering{Arabic}

\begin{abstract}

Learning natural, animal-like locomotion from demonstrations has become a core paradigm in legged robotics.
While motion tracking can reproduce reference gaits, many approaches still require substantial tuning and depend on reference motion inputs at deployment, which can limit responsiveness to task objectives and reduce adaptability.
We present APEX (\underline{A}ction \underline{P}riors enable \underline{E}fficient E\underline{x}ploration), a motion-tracking reinforcement learning (RL) framework that removes deployment-time dependence on reference motion inputs, improves sample efficiency, and reduces tuning effort.
APEX integrates demonstrations into RL via \textit{decaying action priors}, which guide early exploration toward demonstration-consistent actions and then fade to zero, yielding a pure RL policy at deployment.
This is combined with a multi-critic framework that separates style and task + regularization learning signals.
Moreover, APEX enables a single policy to learn diverse motions and transfer reference-like styles across different terrains and velocities, while remaining robust to variations in training parameters.
We validate our method in simulation on both humanoid and quadruped robots, and with zero-shot deployment on a Unitree Go2 robot.
Project Page and code: \url{https://marmotlab.github.io/APEX/}
\end{abstract}


\section{Introduction}\label{sec:intro}

Legged animals exhibit agile and adaptable locomotion in complex environments, inspiring robots to achieve similarly robust and natural behaviors. 
Realizing this goal requires generating natural motion styles, traversing unstructured terrain, and transitioning seamlessly between gaits.

Traditional approaches to legged locomotion rely on optimal control with predefined gaits tailored to specific tasks~\cite{8594448, 7989557, kim2019highly}. While effective in structured environments, these methods face challenges in generalizing to unseen tasks and are computationally intensive. Reinforcement learning (RL) has emerged as a powerful alternative, leveraging full robot dynamics for better generalization, and providing computationally efficient hardware deployment~\cite{Hwangbo2019LearningAA, miki2022learning,margolis2023walk,10161080, he2025asap}. 
However, RL-based methods demand careful reward engineering, particularly since individual motion styles are often difficult to capture with explicit reward terms.

\begin{figure}[]
    \centering
    \includegraphics[width=1.0\linewidth, trim=0cm 0cm 0cm 0cm, clip]{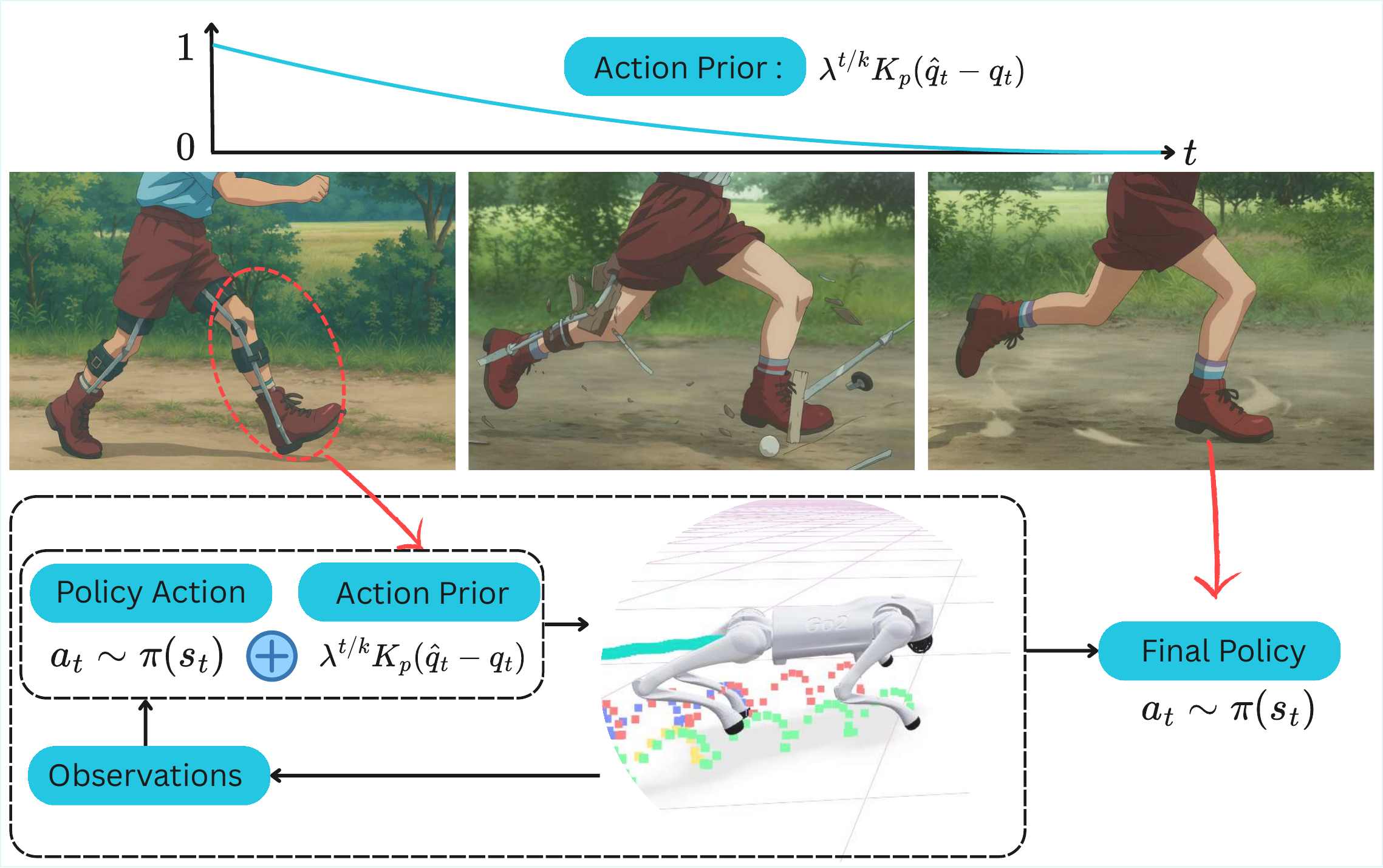}
    \caption{Illustration of APEX's decaying action priors: like braces that stabilize early motion before breaking free, the priors guide exploration at the start of training and fade to zero, enabling a pure RL policy that runs reference-free at deployment. Images inspired by Forrest Gump (1994).}
    \label{fig:gump_metaphor}
\end{figure}

\begin{figure*}
    \centering
    \includegraphics[width=0.95\linewidth, trim=0cm 0cm 0cm 0cm, clip]{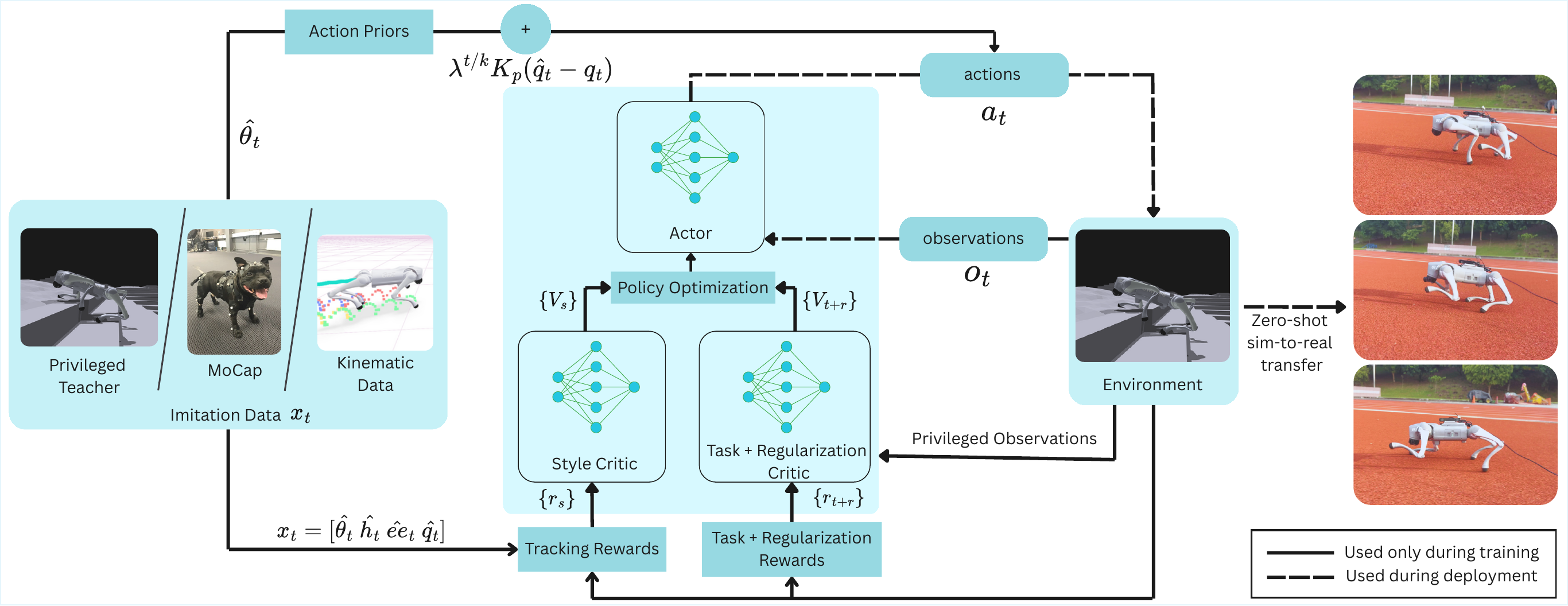}
    \caption{Overview of the APEX framework. Only dashed lines are required during deployment; 1) \textbf{Demonstration data} can be collected from a privileged teacher policy, motion capture data, or other sources such as handcrafted animations; 2) \textbf{Action priors} (Sec.~\ref{subsec:decap}) are calculated from the demonstration's kinematic joint data and added to the policy actions to guide exploration; 3) \textbf{Multi-critic} (Sec.~\ref{subsec:multi-critic}) PPO is used to train the final policy with both style and task+regularization rewards; 4) the trained policy is finally transferred to hardware zero-shot.}
    \label{fig:flowchart}
\end{figure*}

Imitation learning (IL) alleviates these challenges by using demonstrations (e.g., motion capture~\cite{RoboImitationPeng20, zhang2025motion} or video~\cite{peng2018sfv, stamatopoulou2024sds}) to guide policy training. 
Motion tracking based on imitation rewards, such as DeepMimic~\cite{2018-TOG-deepMimic}, remains widely used for its simplicity and effectiveness~\cite{truong2025beyondmimic, he2025asap}. In many practical deployments, however, imitation must be paired with explicit task objectives (e.g., tracking commanded velocities or traversing uneven terrain), where the controller should preserve a desired style while completing the tasks.
Many existing tracking formulations incorporate demonstrations as reference motion inputs (e.g., via phase variables~\cite{he2025asap} or goal poses in the state~\cite{truong2025beyondmimic, RoboImitationPeng20, 2018-TOG-deepMimic}), which can couple the policy to reference timing and reduce flexibility when task demands change.
In addition, reward-based motion tracking often requires careful tuning of the style and task rewards~\cite{xie2026kungfubot}. Because the right balance can vary across motions, terrains, and command distributions, it can be difficult to design a single tracker that both pursues task objectives and preserves the intended motion style.

To address this, we present APEX (Action Priors enable Efficient Exploration), a reinforcement-learning framework for style-aware locomotion control. Rather than treating motion tracking as an end in itself, APEX uses demonstrations to guide exploration \emph{during training} and learns a policy that runs \emph{reference-free} at deployment, reducing reliance on hand-tuned tracking scaffolds.
At the core of APEX are decaying action priors (Fig.~\ref{fig:gump_metaphor})~\cite{Sood2023DecAPD}, which guide early exploration toward demonstration-consistent actions and then fade to zero so that the final controller is a pure RL policy. We complement this with a multi-critic PPO formulation that separates style and task + regularization signals, improving stability when these objectives compete. 
Our contributions are as follows:
\vspace{-2pt}
\begin{itemize}\setlength{\itemsep}{1pt}\setlength{\parsep}{0pt}\setlength{\topsep}{1pt}
\item We learn controllers that track locomotion styles while following task commands (e.g., velocity) \emph{without} deployment-time reference motion or phase inputs.
\item  We show that decaying action priors with multi-critic PPO reduce reward sensitivity and tuning effort, while yielding a robust motion prior.
\item We evaluate on quadrupeds and humanoids in simulation, learn multiple gaits with transitions in a single policy, and demonstrate zero-shot sim-to-real transfer on Unitree Go2 across terrains and velocities~\cite{go2}.
\end{itemize}

Across our evaluated skills, APEX improves reward and convergence relative to reference-free baselines (up to $87\%$ higher reward on dynamic motions) and remains stable under large changes in tracking sensitivity (up to $10\times$) and reward scaling (up to $30\times$).
Together, these results support APEX as a practical and sample-efficient approach to style-aware locomotion tracking.


\section{Background and Related Work}
\label{sec:lit_review}
RL has been widely applied to enable legged robots to reproduce natural locomotion from demonstrations.
Motion tracking approaches, such as DeepMimic~\cite{2018-TOG-deepMimic}, use dense feature-based tracking rewards to reproduce reference trajectories and achieve natural motions.
This reward-based tracking template continues to be used in quadruped locomotion both as a direct baseline for motion transfer~\cite{li2023crossloco} and as an imitation pre-training stage for learning reusable motion priors~\cite{zhang2025motion}.
However, these tracking methods often rely on demonstration data as explicit reference states~\cite{truong2025beyondmimic, RoboImitationPeng20} or phase variables~\cite{he2025asap}, coupling the policy tightly to demonstration timing and limiting adaptability across speeds and terrains.
In quadrupeds, they are also frequently paired with reference state initialization (RSI) and reference generators~\cite{zargarbashi2024metaloco} to stabilize learning.
Moreover, reward-based tracking commonly requires careful reward shaping~\cite{zhang2025add} and curricula~\cite{he2025asap}, which increases engineering effort and can make performance sensitive to reward scaling~\cite{xie2026kungfubot}.

To alleviate the reliance on handcrafted rewards, adversarial imitation methods such as AMP~\cite{Peng_2021, escontrela2022adversarialmotionpriorsmake, yang2024generalized} replaced manually designed objectives with a discriminator trained to distinguish real from generated motions.
While this formulation removes the need for explicit reward shaping, it can suffer from training instability and mode collapse, as is common in GAN training~\cite{zhang2025motion}.
Variants such as ASE~\cite{peng2022ase} and VMP~\cite{Serifi2024VMPVM} mitigate these issues through latent conditioning; however, adversarial instability can remain a practical challenge.
Diffusion-based frameworks (e.g., DiffuseLoco~\cite{huang2024diffuseloco}, BeyondMimic~\cite{truong2025beyondmimic}, FALCON~\cite{he2025falcon}) provide a generative alternative, replacing discriminators with denoising processes that model motion trajectories as iterative refinements of noise.
These often exhibit stronger planning capabilities, but their reliance on multi-step denoising during inference can introduce notable computational overhead~\cite{huang2024diffuseloco, he2025demystifying}.

Recently, a line of research has explored bridging imitation and reinforcement learning through structured guidance. For instance,~\cite{10752370,Smith2023LearningAA} adopt two-stage pipelines, in which policies are first trained through imitation learning and subsequently fine-tuned via reinforcement learning.
Kang et al.~\cite{kang2025learning} distilled demonstration trajectories into compact latent encodings refined through residual learning, while other hierarchical pipelines similarly pre-train motion priors via reward-based tracking and then learn task adaptation policies for rough terrain and navigation~\cite{zhang2025motion}.
Other works are similarly initialized from imitation objectives before fine-tuning for downstream tasks~\cite{truong2025beyondmimic, kalaria2025dreamcontrol, weng2025hdmilearninginteractivehumanoid}, yet they often rely on distinct training phases or persistent access to demonstrations (or reference-generating mechanisms) during deployment.
Our approach, APEX, follows this direction of transitioning from imitation to RL but achieves it in a single, end-to-end training stage.
Further, it removes deployment-time dependence on reference data and reduces tuning effort.


\section{APEX Framework}

We first describe the key components of our approach: action priors and multiple critics, along with the theoretical reasoning behind them. We then discuss important implementation details. Fig.~\ref{fig:flowchart} provides an overview of APEX.


\subsection{Decaying Action Priors}
\label{subsec:decap}

We augment policy actions with a time-decayed prior, building on the DecAP idea introduced by Sood et al.~\cite{Sood2023DecAPD}. While prior work demonstrated improved learning speed for simple torque-based walking with noiseless imitation, we extend the formulation to a general reward-based motion-tracking setting and analyze how the prior shapes exploration and stabilizes policy optimization. DecAP is described as torque-space addition. We instead use the action-space form, which preserves the same intuition while being more general for RL settings where reference actions are available directly in the policy action space. In our setting, this is effectively equivalent, since selected actions are passed through a PD controller before actuation(Fig.~\ref{fig:gump_metaphor}). 

Let the policy sample $a_t \sim \pi_\theta(\cdot \mid s_t)$. From demonstrations, we compute a (deterministic) prior $\beta_t$ and execute
\begin{equation}
\label{equation:exec_actions}
u_t = a_t + c_n \beta_t,\qquad c_n=\lambda^{n/k},
\end{equation}

where $n$ is the global environment step, $\lambda\in(0,1)$, and $k>1$ controls the decay rate. The term $c_n\beta_t$ is the \emph{action prior}. In our setting, demonstrations provide the per-timestep joint angles $q_t^{ref}$, and the action prior is formed in the same action space as the policy (here, joint angles).
Notably, the prior is added directly to the raw RL action, without additional scaling or normalization beyond the decay factor.

This prior can be incorporated into reward-based motion tracking by augmenting the executed action with an additive, time-decayed term. Since, for continuous action spaces, neural policies are typically initialized with weights that yield outputs close to zero, initial actions are small perturbations around demonstration-guided behavior. As the prior decays, the policy must progressively amplify its own contribution and ultimately perform the task independently. Thus, action priors act as an initial nudge toward expert-like regions while allowing unconstrained RL exploration during training.

The prior is intended to guide exploration rather than serve as a high-fidelity controller: even imperfect prior actions can suffice if they guide policy rollouts towards high-performance regions of the state-action space.
We show this in Sec.~\ref{sec:Results}.


\subsubsection{Effect of Action Priors on PPO}

With gradually fading action priors scaffolding the agent's actions, the agent initially interacts with an effectively easier MDP: a shaped version of the original problem that guides rollouts toward higher-quality regions of the state-action space. This setup serves as a form of curriculum learning, where the complexity of the control problem increases as the priors decay.
We hypothesize two main benefits:
(1) \textit{Improved initialization and denser rewards}: the prior places rollouts near higher-imitation-reward regions, stabilizing early behavior and increasing reward density;
(2) \textit{Stabilized PPO updates}: structured executed actions reduce noise in advantage estimation, leading to smoother policy updates.
Additionally, this guidance helps the policy remain robust under heavy domain randomization (DR), where unguided policies may collapse to overly conservative behaviors~\cite{He2024LearningHR, he2025asap}.

At each timestep $t$, the agent samples a policy action $a_t \sim \pi_\theta(\cdot \mid s_t)$. A prior $\beta_t \sim p_\beta(\cdot \mid s_t,t)$ is added based on the demonstration data, where $p_\beta$ is independent of $\theta$. In our implementation, $\beta_t$ is deterministic, but we use stochastic notation for generality. We assume continuous action spaces for the policy action $a_t$, prior $\beta_t$, and executed action $u_t$, though our framework can also be adapted to discrete actions.
The executed action is Eq.~\ref{equation:exec_actions}.

The environment evolves according to the Markovian transition dynamics $\mathcal{P}(s_{t+1} \mid s_t, u_t)$, which is independent of $\theta$, and yields rewards $r(s_t, u_t, s_{t+1})$.
Rollouts are generated by $u_t$, but PPO ratios are computed using $\pi_\theta(a_t\mid s_t)$ since $a_t$ is the policy-sampled random variable.

\emph{Theorem.}
Let $\pi_{\mathrm{APEX}}$ denote the executed policy induced by $\pi_\theta$ and $p_\beta$ under $u_t=a_t+c_n\beta_t$, and let $J(\theta)=\mathbb{E}_{\tau\sim\pi_{\mathrm{APEX}}}[\sum_{t=0}^\infty \gamma^t r_t]$.
If $p_\beta(\cdot \mid s_t,t)$ and $c_n$ are $\theta$-independent, then the likelihood-ratio policy gradient remains unbiased:
\[
\begin{aligned}
\nabla_\theta J(\theta)
&= \mathbb{E}_{\tau \sim \pi_{\mathrm{APEX}}}\!\left[
\sum_{t=0}^\infty
\nabla_\theta \log \pi_\theta(a_t \mid s_t)\right. \\
&\qquad \left.\cdot \big( Q^{\pi_{\mathrm{APEX}}}(s_t, u_t) - b(s_t) \big)
\right],
\end{aligned}
\]
for any baseline $b(s_t)$ independent of $\theta$.

\emph{Proof.}
All $\theta$-dependence enters through $\pi_\theta(a_t\mid s_t)$. Since $p_\beta$, $c_n$, and $\mathcal{P}(s_{t+1}\mid s_t,u_t)$ are $\theta$-independent and $u_t$ is a $\theta$-independent transformation of $(a_t,\beta_t)$, the REINFORCE likelihood-ratio identity holds unchanged~\cite{Williams2004SimpleSG}.


\subsection{Multiple Critics for Task Decomposition}
\label{subsec:multi-critic}

Balancing style imitation with task completion often requires tedious reward-weight tuning~\cite{zhang2025add}. Coupled reward terms can also hurt generalization to different terrains, where the policy might overfit to the imitation data and fail to explore, or learn to complete the task but fail to replicate the style of the imitation data~\cite{wen2025constrained}. To reduce this sensitivity, APEX uses a shared actor with multiple critic heads (two by default), following recent legged-robot literature~\cite{Zargarbashi2024RobotKeyframingLL, liu2025discrete, huang2025learning}.

We split rewards into two groups (Table~\ref{tab:reward_structure}): \textit{group1} (style/imitation) and \textit{group2} (task + regularization). The critic outputs one value per group,
\(V(s)=[V_{\text{style}}(s),\,V_{\text{task}}(s)]\).
We compute Generalized Advantage Estimation (GAE) separately for each reward group, normalize each group's advantage independently, and then form the policy advantage as a weighted sum
\(\hat{A}=\sum_g \alpha_g \tilde{A}_g\)
where we use equal weights for our experiments ($\alpha_g = 0.5$).
The actor is updated using \(\hat{A}\), while the value loss is averaged across critic heads. This decomposition improves credit assignment between ``move like the expert'' and ``solve the task,'' reducing sensitivity to manual reward balancing~\cite{wen2025constrained}.

\begin{table}[tb]
\centering
\scriptsize
\setlength{\tabcolsep}{4.5pt}
\renewcommand{\arraystretch}{1.15}
\caption{Reward terms grouped into two critic heads: task+regularization and style (imitation). 
Tracking rewards use 
$\phi(\mathbf{e}) = \exp\!\left(-{\|\mathbf{e}\|^2}/{\sigma}\right)$, 
where $\mathbf{e}$ is the tracking error and $\sigma$ controls sensitivity.}
\begin{tabular}{l c c c}
\toprule
\textbf{Reward Term} & \textbf{Expression} & \textbf{Weight} & \textbf{$\sigma$} \\
\midrule
\multicolumn{4}{l}{\textit{Task + Regularization (Critic: task)}} \\
\midrule
Linear velocity tracking   & $\phi(\mathbf{v}_t^{\text{cmd}} - \mathbf{v}_t)$ & $1.0$ & $0.3$ \\
Angular velocity tracking  & $\phi(\boldsymbol{\omega}_t^{\text{cmd}} - \boldsymbol{\omega}_t)$ & $0.9$ & $0.25$ \\
Feet slip                  & $\|\mathbf{v}^{\text{foot}}_{xy}\|$              & $-0.04$ & -- \\
Torques                    & $\|\boldsymbol{\tau}\|^2$                         & $-0.0001$ & -- \\
Action rate                & $\|\Delta \mathbf{a}_t\|^2$                       & $-0.01$ & -- \\
Reference height penalty   & $(h_z - h_z^{\text{cmd}})^2$                       & $-10.0$ & -- \\
\midrule
\multicolumn{4}{l}{\textit{Style / Imitation (Critic: style)}} \\
\midrule
Tracking: joint positions       & $\phi(\mathbf{q}_t - \mathbf{q}_t^{\text{ref}})$       & $1.5$ & $0.01$ \\
Tracking: end-effector position & $\phi(\mathbf{p}^{\text{foot}}_{xy} - \mathbf{p}^{\text{ref}}_{xy})$ & $1.5$ & $0.01$ \\
Tracking: base orientation      & $\phi(\mathbf{Q}_t - \mathbf{Q}_t^{\text{ref}})$       & $1.5$ & $0.15$ \\
\bottomrule
\end{tabular}
\label{tab:reward_structure}
\end{table}


\subsection{Learning Multiple Motions in a Single Policy}

To train a single policy over multiple motions, we append a discrete scalar \emph{motion selector} $z$ to the actor's observations. We encode the selector as a normalized ID,
\begin{equation*}
z \in \Big\{ \tfrac{m}{n} \,\Big|\, m = 0,\dots,n-1 \Big\},
\end{equation*}
where $n$ is the number of motions, and each value corresponds to one motion. Conditioning on $z$ allows a single network to represent multiple behaviors without architectural changes beyond this additional input. 
At deployment, switching gaits reduces to setting $z$ to the desired value.
Despite demonstrations containing only steady-state gaits (no transitions) at fixed speeds, the selector-conditioned policy learns direct gait transitions and generalizes each gait across a wide range of commanded velocities.
During training, we sample motions uniformly to ensure balanced coverage across selectors.
We train with four gaits due to limited quadruped motion data; scaling to additional gaits simply increases $n$ (and thus the set of selector values). 
Empirically, APEX's guided exploration improves stability in this multi-gait setting by guiding early rollouts toward the demonstration associated with the selected $z$, which makes learning of the conditional mapping easier.


\subsection{Implementation Details}

\subsubsection{APEX Training}
For all quadruped experiments, we use $\lambda = 0.99$ and $k = 100$ for the action-prior decay. Although this introduces two hyperparameters, we found these values to work consistently across motions, including multi-gait policies. Combined with APEX's robustness to reward scale and sensitivity (Sec.~\ref{subsec:sigma}), this substantially reduces tuning effort.
All policies are trained on a single NVIDIA RTX $4090$ GPU ($24~GB$). Unless otherwise stated, reference-free APEX policies (including the multi-gait policy) are trained for at most $2000$ iterations (approximately $30$ minutes of wall-clock time).

\subsubsection{Observation and Action Space}
The actor observes the gravity vector, joint positions, joint velocities, and a one-step history of actions; it does not require stacked observation histories. The action space matches the environment interface and is executed through a low-level PD controller (i.e., actions specify PD targets). 
The critic additionally receives privileged information: base linear/angular velocities and time-aligned demonstration features at each step. Unlike motion-tracking baselines that condition the actor on phase variables or reference states~\cite{2018-TOG-deepMimic}, we intentionally exclude reference motion inputs from the actor to emphasize reference-free deployment.

\subsubsection{Domain Randomization}
\begin{figure*}
    \centering
    \includegraphics[width=1.0\linewidth, trim=0cm 0cm 0cm 0cm, clip]{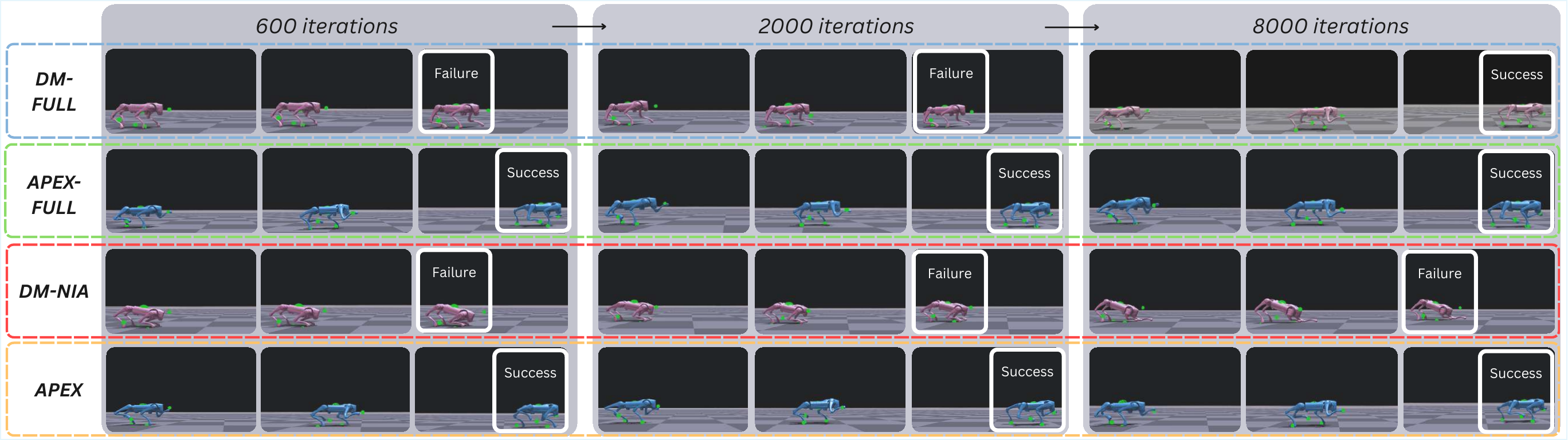}
    \caption{Visual comparison of task tracking across training iterations for DeepMimic (DM) and APEX. Green spheres show body relative reference foot positions.
    Failure here refers to large deviations from either of the tracking/task objectives.
    The APEX policies (blue) demonstrate accurate task tracking from the early stages, highlighting superior sample efficiency even relative to DM-Full (pink). By $2000$ iterations, both APEX-Full and APEX have effectively converged, whereas DM-NIA has converged to a suboptimal solution (as seen in the final pose).}
    \label{fig:compare}
\end{figure*}

We apply domain randomization to improve robustness to the sim-to-real gap and external disturbances. We randomize ground friction in $[0.3, 1.25]$, vary base mass by $\pm 1~\mathrm{kg}$, scale link masses by $\pm 10\%$, perturb the center of mass by up to $(0.05, 0.03, 0.03)~\mathrm{m}$, and randomize PD gains in $[0.9, 1.1]$. Random pushes are applied every $4$--$5~\mathrm{s}$ (linear velocity up to $0.4~\mathrm{m/s}$, angular velocity up to $0.6~\mathrm{rad/s}$). In practice, APEX remains stable under stronger domain randomization without collapsing to overly conservative behaviors.


\section{Results}
\label{sec:Results}

We evaluate APEX on style-aware locomotion, where the controller must (i) preserve a demonstrated gait style and (ii) satisfy task objectives such as commanded velocity tracking. We compare against variants of DeepMimic~\cite{2018-TOG-deepMimic} mentioned in Table~\ref{tab:variant-comparison}.
DeepMimic-style tracking remains a strong and commonly used baseline for precise reference tracking (Sec.~\ref{sec:lit_review}); for example, the recent MimicKit benchmark reports that motion tracking methods such as DeepMimic achieve lower tracking errors than distribution-matching objectives (e.g., AMP) across diverse motions~\cite{MimicKitPeng2025}.
For evaluations, action priors are set to zero, so results reflect the learned policy at reference-free deployment. We report tracking performance via RMSE and mean reward (Table~\ref{tab:errors-symbols}) and analyze sample efficiency and robustness using learning curves and reward-scale/sensitivity sweeps (Figs.~\ref{fig:compare} and~\ref{fig:sigma}). We study the following questions:
\begin{itemize}
\item \textit{Q1: Tracking \& sample efficiency in reference-free deployment.} How does APEX compare to DeepMimic variants (DM-Full and the runtime reference-free ablation DM-NIA) in tracking accuracy and sample efficiency across diverse locomotion skills?
\item \textit{Q2: Generalization \& sim-to-real.} How does removing explicit reference/phase inputs affect generalization beyond the demonstrations across terrains and commanded velocities, and to what extent does APEX transfer from simulation to the real robot?
\item \textit{Q3: Reward robustness \& ablations.} How robust is APEX to reward scaling and sensitivity, and how do the multi-critic architecture and decaying action priors affect convergence and stability?
\item \textit{Q4: High-DoF robustness.} Does APEX remain effective on high-DoF humanoid motion tracking, where action-prior guidance is weaker than in quadrupeds?
\end{itemize}


\subsection{Reference-Free Deployment Performance}
\label{motion-tracking-results}
To evaluate APEX’s core claim: removing deployment-time motion inputs without sacrificing performance, we use reward-based motion tracking as a representative setting. This setting presents the key trade-off in style-aware locomotion: matching a demonstrated gait style while following task commands (e.g., velocity). We therefore report \emph{style} errors ($q$, $x_{ee}$), \emph{task} errors ($v$, $h$), and mean reward (Table~\ref{tab:errors-symbols}). Results are grouped into \emph{Full} (DM-Full vs.\ APEX-Full), which uses the standard reference-conditioned tracking, and \emph{reference-free} (DM-NIA vs.\ APEX), which removes runtime reference/phase inputs as summarized in Table~\ref{tab:variant-comparison}. 
Demonstrations are obtained from retargeted animal motion capture~\cite{RoboImitationPeng20} and additional locomotion datasets~\cite{margolis2023walk, yoon2025spatio}, showing that APEX is not tied to a single data source.

\begin{table}[tb]
\centering
\small
\renewcommand{\arraystretch}{1.1}
\setlength{\tabcolsep}{3pt}
\caption{Deployment-time dependencies across policy variants. 
\req = required, \free = not required. 
All variants use privileged imitation features in the critic during training.
}
\begin{tabular}{lcccc}
\toprule
{Requires} & {DM-Full} & {APEX-Full} & {DM-NIA} & {APEX} \\
\midrule
Runtime ref. motions   & \req & \req & \free  & \free \\
Phase variable         & \req & \req & \req  & \free \\
Ref. state init.       & \req & \req & \req  & \free \\
\bottomrule
\end{tabular}
\label{tab:variant-comparison}
\end{table}

\begin{table}[!htbp]
\centering
\small
\setlength{\tabcolsep}{4pt}
\renewcommand{\arraystretch}{1.05}
\caption{RMSE across gaits.
Metrics: $v$ = velocity [m/s], $h$ = height [m], $q$ = joint position [rad], $x_{ee}$ = end-effector position [m], and $R$ = mean reward. Best within Full (DM-Full vs APEX-Full) is shaded green; best within reference-free (DM-NIA vs APEX) is shaded light green. Impr. reports relative mean-reward gain of APEX.
}
\begin{tabular}{l l S S !{\vrule width 0.2pt} S S}
\toprule
Skill & Metric & {DM-Full} & {APEX-Full} & {DM-NIA} & {APEX} \\
 & & {\scriptsize (10k iters)} & {\scriptsize (10k iters)} & {\scriptsize ($\geq$10k iters)} & {\scriptsize (2k iters)} \\
\midrule
\multirow{5}{*}{Canter}
  & $q$      & 0.078 & \cellcolor{best}{0.077} & 0.293 & \cellcolor{secondbest}{0.144} \\
  & $h$      & \cellcolor{best}{0.012} & \cellcolor{best}{0.012} & 0.025 & \cellcolor{secondbest}{0.019} \\
  & $x_{ee}$ & 0.061 & \cellcolor{best}{0.060} & 0.076 & \cellcolor{secondbest}{0.072} \\
  & $v$      & 0.146 & \cellcolor{best}{0.143} & 1.14  & \cellcolor{secondbest}{0.161} \\
  & $R$      & 35.67 & \cellcolor{best}{36.08} & 18.43 & \cellcolor{secondbest}{31.94} \\
  \cmidrule(lr){2-6}
  & Impr. & \multicolumn{2}{c}{\textbf{{+0.87\%}}} & \multicolumn{2}{c}{\textbf{{+73.3\%}}} \\
\midrule
\multirow{5}{*}{Trot}
  & $q$      & 0.097 & \cellcolor{best}{0.039} & 0.979 & \cellcolor{secondbest}{0.077} \\
  & $h$      & 0.005 & \cellcolor{best}{0.004} & 0.012 & \cellcolor{secondbest}{0.007} \\
  & $x_{ee}$ & 0.047 & \cellcolor{best}{0.043} & 0.137 & \cellcolor{secondbest}{0.041} \\
  & $v$      & 0.098 & \cellcolor{best}{0.056} & 1.065 & \cellcolor{secondbest}{0.064} \\
  & $R$      & 40.2  & \cellcolor{best}{41.5}  & 18.87 & \cellcolor{secondbest}{35.37} \\
  \cmidrule(lr){2-6}
  & Impr. & \multicolumn{2}{c}{\textbf{{+3.2\%}}} & \multicolumn{2}{c}{\textbf{{+87.5\%}}} \\
\midrule
\multirow{5}{*}{Hopturn}
  & $q$      & 0.070 & \cellcolor{best}{0.057} & 0.157 & \cellcolor{secondbest}{0.101} \\
  & $h$      & 0.049 & \cellcolor{best}{0.023} & 0.034 & \cellcolor{secondbest}{0.027} \\
  & $x_{ee}$ & \cellcolor{best}{0.016} & \cellcolor{best}{0.016} & 0.030 & \cellcolor{secondbest}{0.021} \\
  & $v$      & \cellcolor{best}{0.060} & 0.066 & \cellcolor{secondbest}{0.066} & 0.071 \\
  & $R$      & 26.96 & \cellcolor{best}{28.48} & 21.08 & \cellcolor{secondbest}{26.08} \\
  \cmidrule(lr){2-6}
  & Impr. & \multicolumn{2}{c}{\textbf{{+5.6\%}}} & \multicolumn{2}{c}{\textbf{{+23.7\%}}} \\
\midrule
\multirow{5}{*}{Pronk}
  & $q$      & 0.102 & \cellcolor{best}{0.071} & 0.123 & \cellcolor{secondbest}{0.091} \\
  & $h$      & 0.038 & \cellcolor{best}{0.014} & 0.015 & \cellcolor{secondbest}{0.013} \\
  & $x_{ee}$ & 0.040 & \cellcolor{best}{0.035} & 0.047 & \cellcolor{secondbest}{0.040} \\
  & $v$      & 0.083 & \cellcolor{best}{0.081} & \cellcolor{secondbest}{0.091} & 0.098 \\
  & $R$      & 39.14 & \cellcolor{best}{40.87} & 34.25 & \cellcolor{secondbest}{37.10} \\
  \cmidrule(lr){2-6}
  & Impr. & \multicolumn{2}{c}{\textbf{{+4.4\%}}} & \multicolumn{2}{c}{\textbf{{+8.3\%}}} \\
\midrule
\multirow{5}{*}{Sidestep}
  & $q$      & 0.042 & \cellcolor{best}{0.041} & 0.171 & \cellcolor{secondbest}{0.099} \\
  & $h$      & 0.018 & \cellcolor{best}{0.017} & 0.021 & \cellcolor{secondbest}{0.018} \\
  & $x_{ee}$ & \cellcolor{best}{0.009} & \cellcolor{best}{0.009} & 0.035 & \cellcolor{secondbest}{0.022} \\
  & $v$      & 0.146 & \cellcolor{best}{0.143} & \cellcolor{secondbest}{0.178} & 0.206 \\
  & $R$      & 30.91 & \cellcolor{best}{31.54} & 24.59 & \cellcolor{secondbest}{26.55} \\
  \cmidrule(lr){2-6}
  & Impr. & \multicolumn{2}{c}{\textbf{{+2\%}}} & \multicolumn{2}{c}{\textbf{{+7.97\%}}} \\
\midrule
\multirow{5}{*}{Pace}
  & $q$      & 0.065 & \cellcolor{best}{0.047} & 0.344 & \cellcolor{secondbest}{0.113} \\
  & $h$      & 0.014 & \cellcolor{best}{0.006} & 0.027 & \cellcolor{secondbest}{0.015} \\
  & $x_{ee}$ & 0.038 & \cellcolor{best}{0.033} & 0.088 & \cellcolor{secondbest}{0.058} \\
  & $v$      & 0.124 & \cellcolor{best}{0.047} & 0.420 & \cellcolor{secondbest}{0.055} \\
  & $R$      & 39.58 & \cellcolor{best}{42.10} & 34.97 & \cellcolor{secondbest}{35.66} \\
  \cmidrule(lr){2-6}
  & Impr. & \multicolumn{2}{c}{\textbf{{+6.4\%}}} & \multicolumn{2}{c}{\textbf{{+2.0\%}}} \\
\bottomrule
\end{tabular}
\label{tab:errors-symbols}
\end{table}

\textbf{Baselines.} DM-Full follows standard DeepMimic design choices, including reference state initialization, a phase variable, and imitation references in the actor observations~\cite{2018-TOG-deepMimic}.
This choice reflects current practice in motion imitation benchmarks (e.g., MimicKit), where DeepMimic-style tracking remains a strong baseline when accurate per-clip replication is desired~\cite{MimicKitPeng2025}.
DM-NIA removes imitation references from the actor while retaining other tracking optimizations (phase and reference state initialization), isolating the effect of removing deployment-time imitation inputs for reference-dependent trackers.

\textbf{APEX variants.} APEX-Full retains DM-Full's design choices for direct comparison, while APEX removes deployment-time motion inputs by excluding runtime imitation references, phase variables, and reference state initialization from the actor (Table~\ref{tab:variant-comparison}). All variants are trained without position-error terminations.

\subsubsection{Quadrupeds}
Fig.~\ref{fig:compare} compares training qualitatively: APEX achieves accurate tracking from early iterations and converges within $2000$ iterations, while DM-Full converges more slowly and DM-NIA struggles once runtime imitation inputs are removed. Quantitatively, Table~\ref{tab:errors-symbols} shows that APEX preserves strong performance while removing deployment-time reference/phase inputs.
The gap is most pronounced for highly dynamic skills (e.g., canter and trot), where reference-free baselines often collapse to conservative behaviors under domain randomization. In these regimes, APEX helps the most, yielding denser style rewards and enabling PPO to discover stable, highly dynamic gaits that are difficult to reach via unguided exploration. Notably, the small reward drop in APEX compared to APEX-Full does not indicate degraded style, as evidenced by the qualitative rollouts (Fig.~\ref{fig:compare}) and our accompanying video.

\subsubsection{Humanoids}
\begin{figure}[]
    \centering
    \includegraphics[width=0.49\linewidth]{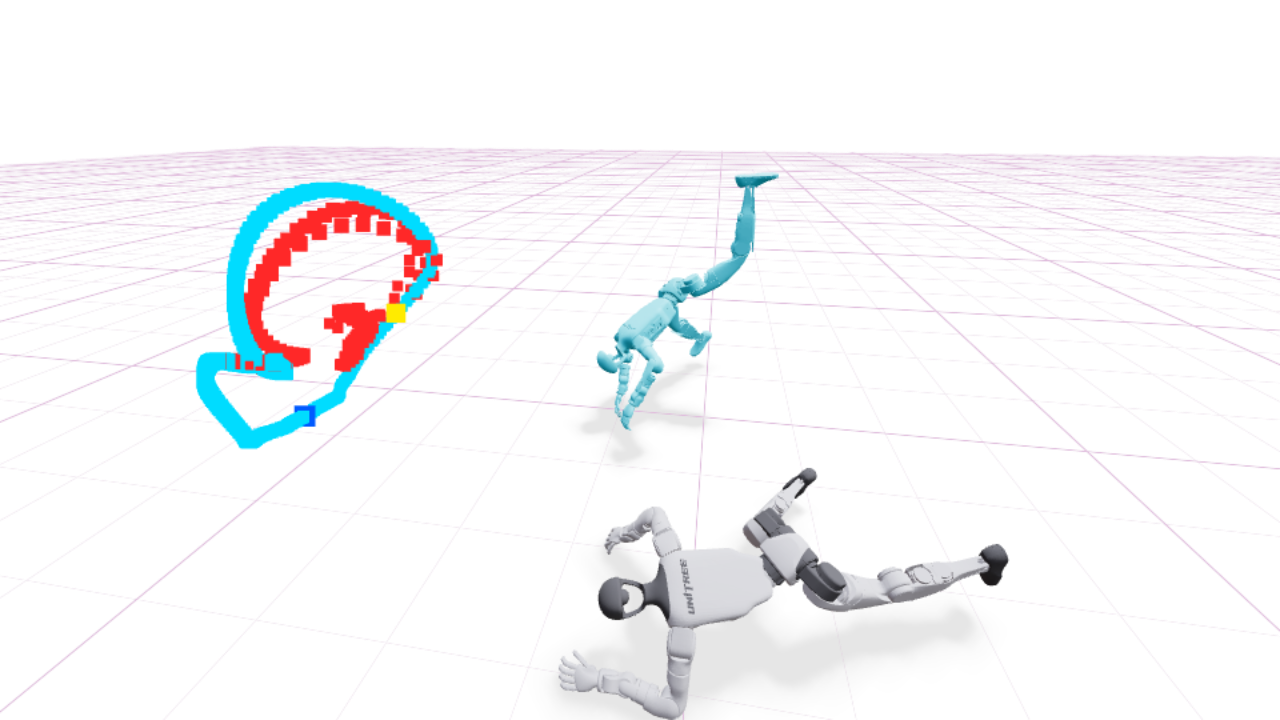}\hfill
    \includegraphics[width=0.49\linewidth]{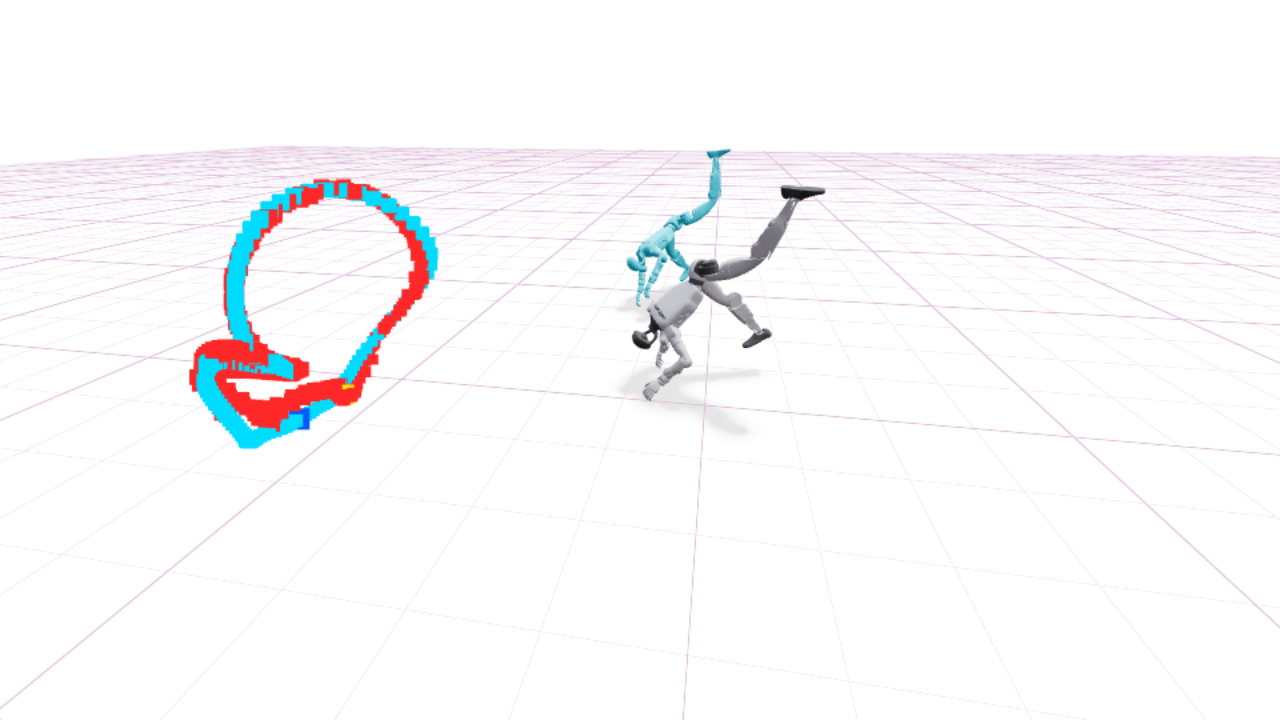}
    \caption{Cartwheel tracking. The blue robot is the reference motion. Curves show the reference wrist path (cyan) and the learned policy (red) in the root-local frame. APEX (right) tracks more closely and lands stably, whereas DM-NIA shows larger drift and an unstable landing (left).}
    \label{fig:cartwheel_spinkick}
\end{figure}

While our main focus is style-aware locomotion with task objectives, we also evaluate APEX on two highly dynamic humanoid motion tracking tasks to verify that our approach does not break down in high-DoF settings. We use MimicKit~\cite{MimicKitPeng2025} framework on the 29-DoF Unitree G1, where action-prior guidance is weaker than in quadrupeds: executing the prior alone does not yield a meaningful tracker, whereas quadrupeds can remain partially aligned to the reference under prior-only control for simpler motions. Table~\ref{tab:humanoid_pd_ablation} shows that APEX remains effective in this regime: even when action-prior guidance cannot faithfully track the motion, the decaying action prior still provides structured early exploration that enables learning. Fig.~\ref{fig:cartwheel_spinkick} illustrates this qualitatively (cartwheel/spin-kick): APEX stays closer to the reference end-effector trajectory, while DM-NIA exhibits larger drift and unstable recovery. Humanoid training runs for $400$M environment timesteps; the action-prior term decays by $\sim 300$M timesteps, after which optimization proceeds as pure RL while retaining the style discovered during the guided phase.

\begin{table}[tb]
\centering
\small
\setlength{\tabcolsep}{4pt}
\renewcommand{\arraystretch}{1.05}
\caption{Humanoid Motion Tracking. $x_{\mathrm{key}}$ denotes key-body position RMSE [m] over \{left/right ankle, head, left/right wrist\}; $q$, $h$, and $v$ are defined as in Table~\ref{tab:errors-symbols}.}
\begin{tabular}{l l S S !{\vrule width 0.2pt} S S}
\toprule
Skill & Metric & {DM-Full} & {APEX-Full} & {DM-NIA} & {APEX} \\
\midrule
\multirow{4}{*}{G1 Spinkick}
  & $q$                & \cellcolor{best}{0.122} & 0.126 & 0.231 & \cellcolor{secondbest}{0.198} \\
  & $h$                & 0.041 & \cellcolor{best}{0.037} & 0.158 & \cellcolor{secondbest}{0.081} \\
  & $x_{\mathrm{key}}$ & 0.071 & \cellcolor{best}{0.065} & 0.264 & \cellcolor{secondbest}{0.167} \\
  & $v$                & 0.328 & \cellcolor{best}{0.311} & 0.620 & \cellcolor{secondbest}{0.450} \\
\cmidrule(lr){2-6}
\multirow{4}{*}{G1 Cartwheel}
  & $q$                & 0.123 & \cellcolor{best}{0.122} & 0.461 & \cellcolor{secondbest}{0.265} \\
  & $h$                & \cellcolor{best}{0.034} & 0.037 & 0.434 & \cellcolor{secondbest}{0.038} \\
  & $x_{\mathrm{key}}$ & 0.128 & \cellcolor{best}{0.113} & 0.806 & \cellcolor{secondbest}{0.272} \\
  & $v$                & 0.341 & \cellcolor{best}{0.329} & 0.505 & \cellcolor{secondbest}{0.347} \\
\bottomrule
\end{tabular}
\label{tab:humanoid_pd_ablation}
\end{table}

\begin{figure*}
    \centering
    \includegraphics[width=0.9\linewidth,   trim=4cm 0cm 4cm 0cm, clip]{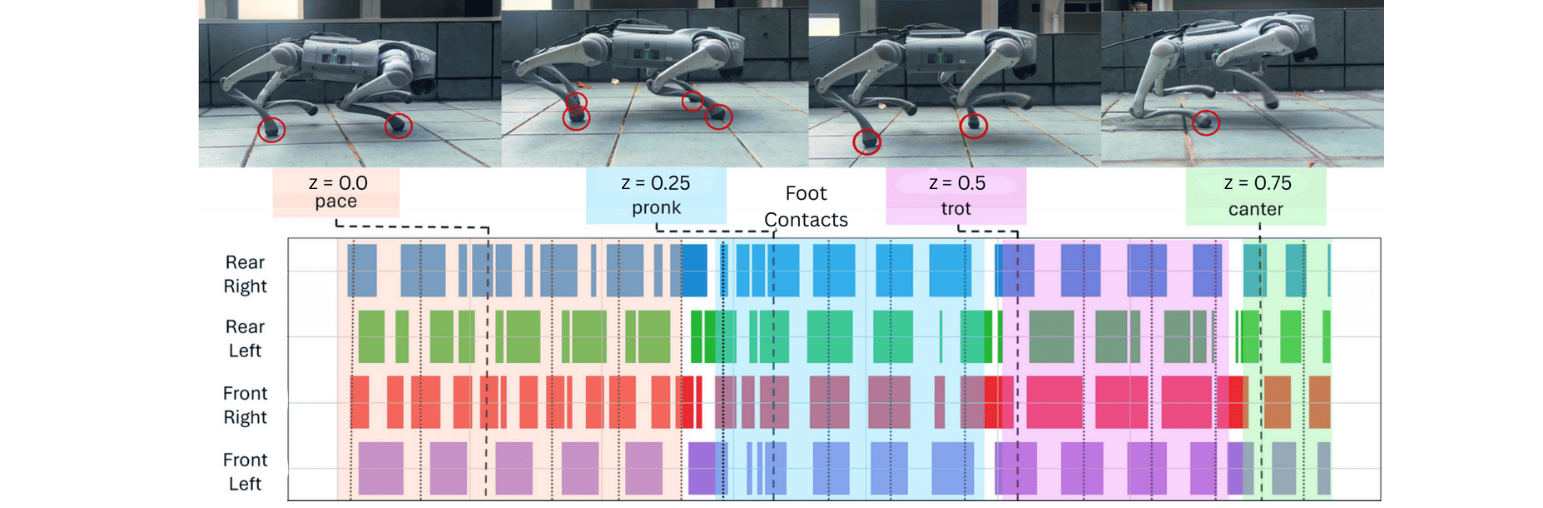}
    \caption{Visualization of gait patterns learned by the APEX multi-gait policy, conditioned on the scalar motion selector $z$. Red circles mark foot–ground contacts in hardware rollouts, while the contact diagrams below illustrate corresponding stance and swing phases across legs. By varying $z$, the policy smoothly transitions between pace ($z=0.0$), pronk ($z=0.25$), trot ($z=0.5$), and canter ($z=0.75$), demonstrating lightweight and scalable gait switching within a single network.}
    \label{fig:APEX_gait_change}
\end{figure*}


\subsection{Reward Sensitivity Analysis}
\label{subsec:sigma}

\begin{figure}[!htbp]
    \centering
    \includegraphics[width=1.0\linewidth]{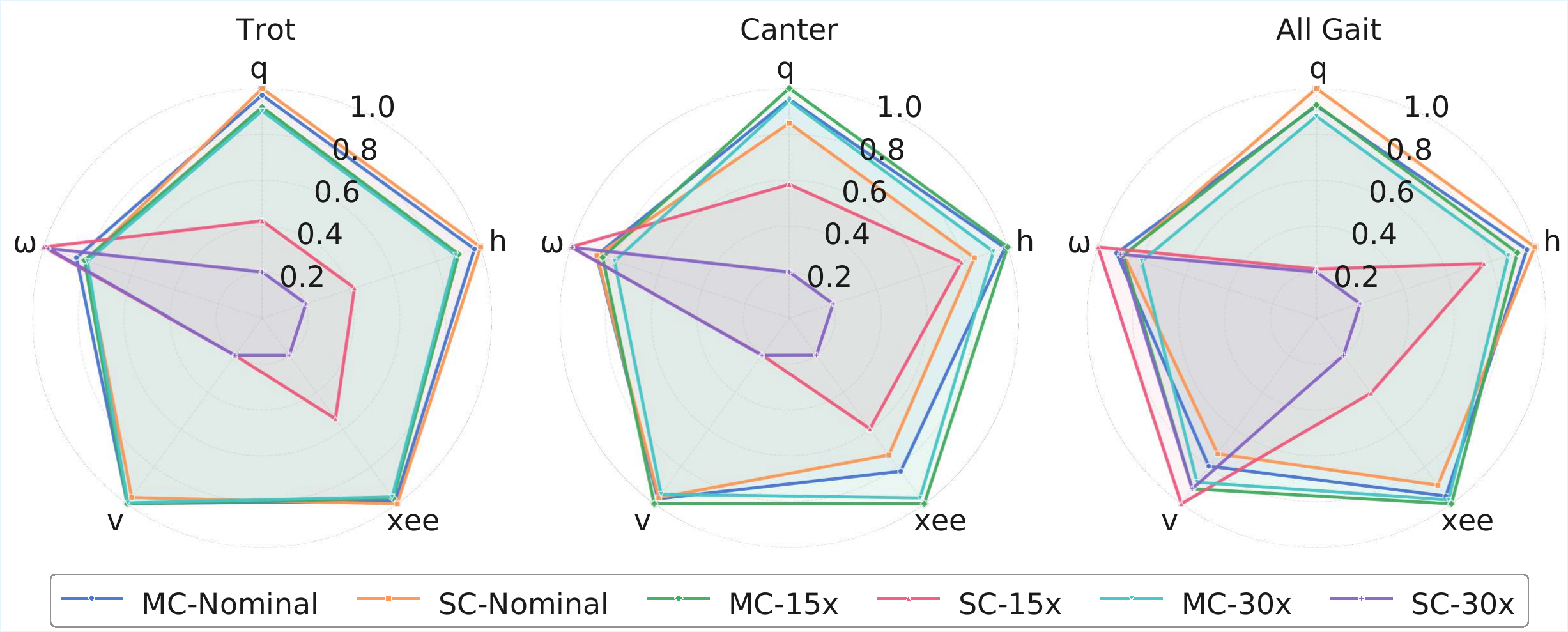}
    \caption{Multi-critic (MC) variants maintain high tracking accuracy across reward scaling, whereas single-critic (SC) variants degrade when velocity-tracking rewards dominate.}
    \label{fig:multi_vs_single_critic}
\end{figure}

\begin{figure}[!tb]
    \centering
    \includegraphics[width=1\linewidth]{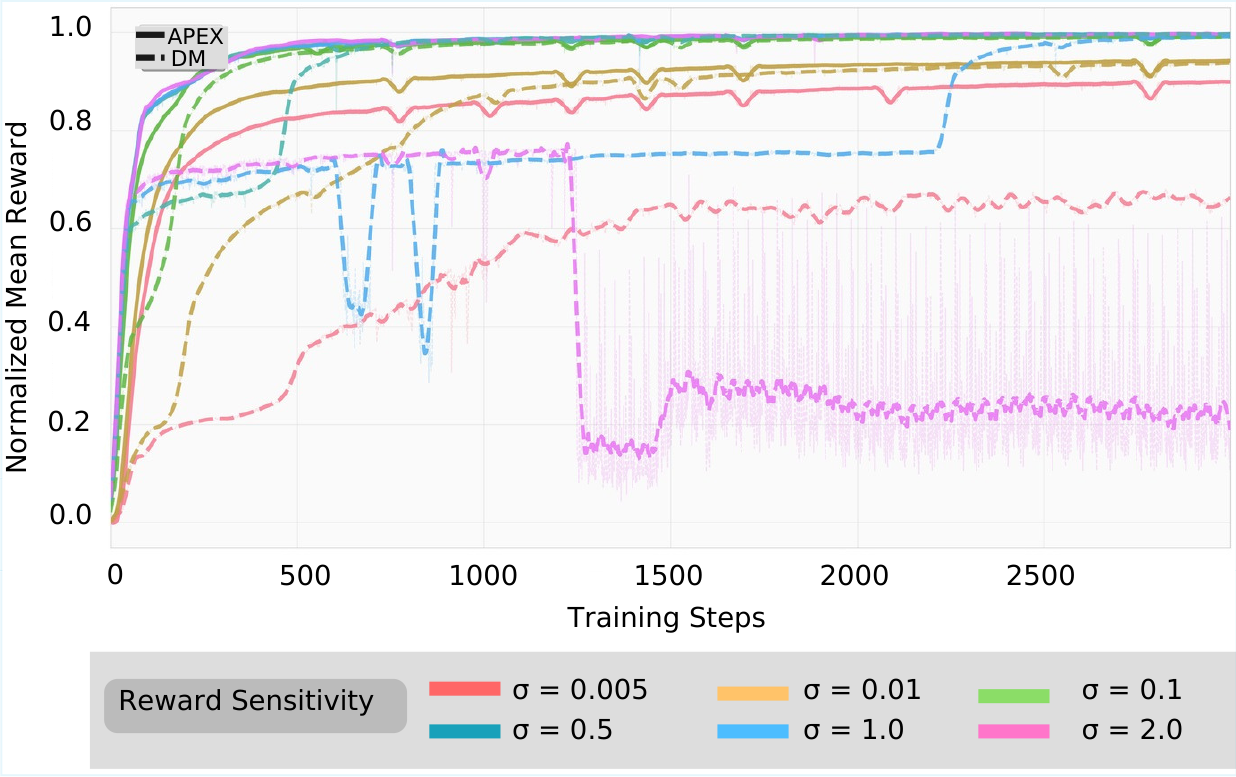}
        \caption{Training performance across reward sensitivities $\sigma$ for APEX-Full and DM-Full. APEX-Full (solid) maintains stable convergence and high rewards across all tested $\sigma$, while DM-Full (dashed) shows instability and drop-off. Similar trends are observed for APEX, with APEX variants converging faster than the DM baselines in our settings.}
    \label{fig:sigma}
\end{figure}

We aim to demonstrate that APEX is less sensitive to reward scaling and tuning hyperparameters.
To this end, Fig.~\ref{fig:multi_vs_single_critic} isolates the effect of multi-critic (MC) learning under varying reward scales (Q3). 
For visualization, Fig.~\ref{fig:multi_vs_single_critic} reports \emph{inverted, normalized RMSE} (i.e., larger values indicate lower error / better tracking).
MC variants remain stable across metrics even when task rewards dominate, while single-critic (SC) variants degrade sharply, most notably under strong velocity-tracking rewards. Fig.~\ref{fig:sigma} further shows that APEX-Full maintains stable convergence across wide variations in the tracking-sensitivity parameter $\sigma$, whereas DM-Full exhibits large fluctuations in both convergence behavior and final reward.

Together, these results suggest a complementary interaction: action priors provide structured early exploration near the demonstrated motion manifold, while the multi-critic decomposition preserves consistent learning signals for style and task objectives as the priors vanish, improving reliability and reducing reward-tuning effort.

\subsection{Multi-Gait Policy on Hardware}
 
As APEX operates without motion input dependency at deployment, we evaluate whether a single policy can encode and seamlessly switch between multiple gaits in a fully reference-free setting by deploying the learned multi-gait policy on Unitree Go2 (Q2).
On hardware, the policy produces distinct and repeatable gait patterns, where each selector value $z \in \{m/n \mid m=0,\dots,n-1\}$ corresponds to a specific gait (pace $z=0$, pronk $z=0.25$, trot $z=0.5$, and canter $z=0.75$). As shown in Fig.~\ref{fig:APEX_gait_change}, these gaits exhibit clear and consistent footfall timings, demonstrating that a single selector-conditioned policy can represent diverse motions while remaining stable at deployment.
A key advantage of this formulation is flexibility: the same gait can be executed across a wide range of commanded velocities, and the robot can transition directly between gaits (e.g., canter to pace) without requiring intermediate gaits.


\subsection{Style Transfer to Terrains and Velocities on Hardware}

\begin{figure}[!htbp]
    \centering
    \includegraphics[width=1.0\columnwidth]{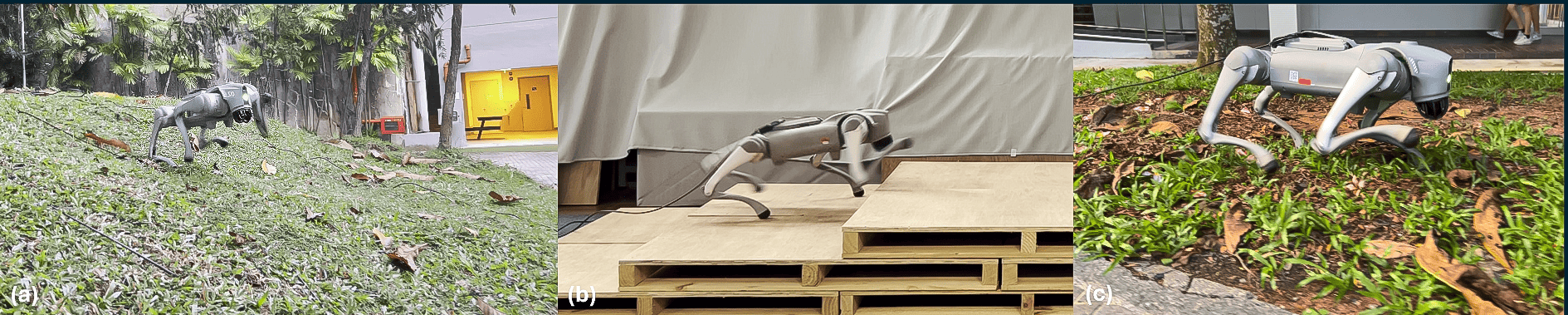}
    \caption{Hardware deployment on unseen terrains: (a) canter on steep uneven slopes ($\sim 30^\circ$), (b) trot over $10$\,cm stairs, and (c) outdoor uneven-terrain robustness (pace shown). All hardware results are in our accompanying video.}
    \label{fig:APEX_uneven_terrain}
\end{figure}

We evaluate APEX policies on real-world terrains not present in the imitation data (Fig.~\ref{fig:APEX_uneven_terrain}). Using only flat-ground demonstrations and training with a rough-terrain curriculum similar to~\cite{rudin2022learningwalkminutesusing}, the learned policies transfer effectively to challenging real world settings: (1) cantering up and down a $\sim 30^\circ$ slope, (2) trotting over $10\,\text{cm}$ stairs ($5/5$ successes), and (3) stable locomotion in unstructured outdoor terrain.

Beyond terrain variation, the learned styles generalize across a wide range of commanded linear and angular velocities, despite the imitation data being collected at fixed gait-specific speeds. DM-Full can also generalize to different commanded velocities, but requires deployment-time reference/phase inputs. Our goal is reference-free deployment while retaining style: DM-NIA, which removes imitation inputs from the actor, fails to preserve the demonstrated gait style. In contrast, APEX maintains style while generalizing across terrains and commands, converging within $2000$ iterations under the same rough-terrain curriculum (Q2).

\section{Conclusion}

We presented APEX, a reinforcement learning framework that combines decaying action priors with a multi-critic architecture for style-aware locomotion with task objectives. Action priors guide early exploration and later decay to zero, yielding a policy that runs reference-free at deployment while retaining the discovered motion style. Across simulated skills, APEX improves sample efficiency and remains robust to large changes in reward scaling and tracking sensitivity. We further show that APEX remains effective on high-DoF humanoid motions, where action-prior guidance is weaker than in quadrupeds. On Unitree Go2, a single multi-gait policy reproduces multiple gaits and transitions directly between them under diverse commands.

Several directions remain for future work. Our current discrete skill selector assumes predefined motion labels; extending it to continuous or unsupervised skill representations could enable smoother blending across behaviors. Incorporating visual feedback for terrain perception could further improve real-world robustness. Finally, adaptive prior decay schedules and transfer to broader domains (e.g., loco-manipulation) may further expand APEX's applicability. \textit{Acknowledgment:} We used ChatGPT for part of the image creation in Fig~\ref{fig:gump_metaphor}, and language polishing in the Introduction and Conclusion; all outputs were critically reviewed and revised by the authors.


\balance

\bibliographystyle{IEEEtran}
\bibliography{citations}

\end{document}